\documentclass[11pt]{article}

\usepackage[preprint]{acl}

\usepackage{times}
\usepackage{latexsym}

\usepackage[T1]{fontenc}
\usepackage[utf8]{inputenc}

\usepackage{microtype}

\usepackage{inconsolata}

\usepackage{graphicx}
\usepackage{subcaption}
\usepackage{booktabs}

\usepackage{amsmath}
\usepackage{amssymb}
\usepackage{amsthm}
\usepackage{enumitem}
\newtheorem{property}{Property}

\title{Hidden Anchors in Multi-Agent LLM Deliberation}

\author{Apurba Pokharel \and Ram Dantu \\
        Department of CSE \\ University of North Texas \\  76207, Denton, TX, USA \\
        \texttt{apurba.pokharel@unt.edu} \and \texttt{ram.dantu@unt.edu}}

\begin{document}
\maketitle
\begin{abstract}
% OLD ABSTRACT (commented out, kept for reference):
% Multi-agent LLM deliberation is treated almost entirely as a black box that
% improves accuracy, with no account of the dynamics that drive it. We model
% deliberation as a closed-loop dynamical system in which each agent carries a
% hidden per-agent anchor, an internal belief that continually pulls its opinion
% regardless of its neighbours. The anchor is identifiable from observed
% trajectories alone. It explains an anomaly that classical linear consensus
% rules such as DeGroot and Friedkin--Johnsen provably forbid. The gold-class
% probability can leave the convex hull of the agents' initial beliefs. Fitting
% the model and validating it on held-out runs turns the procedure into a model
% selection test that detects when a hidden anchor governs deliberation. Across
% three open-weight families the result is a spectrum rather than an all-or
% nothing effect. The mechanism is uniform, every family anchors with
% comparable strength, but the families differ in where the anchor sits. Only
% when the anchor lies far from the initial opinions does deliberation escape
% the hull and require the full closed-loop model.
  Multi-agent LLM deliberation, where agents exchange and revise answers over
  several rounds, is increasingly used to improve reasoning and accuracy, yet
  how and why it works is rarely modelled. Such deliberation mirrors how humans
  reach decisions. As social animals we are pulled both by the group, the herd
  effect that classical opinion-dynamics models such as DeGroot and
  Friedkin--Johnsen capture, and by our own internal belief, which they do not.
  We model multi-agent deliberation as a closed-loop dynamical system in which
  each agent carries a hidden internal belief, its \textbf{anchor}, that
  continually pulls its opinion regardless of its neighbours. We show this
  anchor can be recovered from the deliberation alone, and that it explains a
  behaviour classical consensus rules forbid: an agent's confidence in the
  correct answer can climb past where any agent started, escaping the space (convex
  hull) formed by the initial beliefs. Checking whether the recovered anchor also
  predicts held-out runs (generalizes) gives a simple test for when a model is truly driven by
  such an anchor. Across three open-weight model families this is a spectrum,
  not all-or-nothing. All anchors' influence are about equally strongly, but they differ in
  where the anchor sits, and only when it sits far from the initial opinions
  does deliberation escape the hull and need the full closed-loop model.
\end{abstract}

\section{Introduction}
\textbf{Multi-Agent Systems (MAS):} LLM-based multi-agent systems are showing remarkable performance, and as a
  result their adoption is growing rapidly~\cite{beckerSurvey}.
In particular, deliberation-based multi-agent systems, where agents exchange
  and revise answers over multiple rounds, have seen massive research
  effort~\cite{TODO-du-debate, TODO-liang, TODO-chateval}.
These works are overwhelmingly directed at improving performance by
  designing better deliberation frameworks, while how and why
  deliberation works remains largely unexamined.
Prior work treats deliberation as a black box that empirically improves
 accuracy, and to our knowledge nobody models the deliberation 
  itself as a dynamical system.
This is the focus of our work, and we show that this helps explain the behavior of argentic deliberation.

\textbf{Modelling MAS as a dynamic system:} The opinion-dynamics literature
  offers classical linear consensus rules, DeGroot~\cite{TODO-degroot},
  Hegselmann--Krause~\cite{TODO-hk}, and Friedkin--Johnsen~\cite{TODO-fj}
  (\S\ref{sec:opinion}), but these cannot reproduce LLM deliberation.
% We frequently observe that in LLM deliberation across deliberation runs, the probability of the class that the group believes to be correct
%   rises where as the other probabilities dip.
% In other words, the correct trajectories escape the space (convex hull) formed by the group's initial beliefs.
We frequently observe that across deliberation runs the probability of the
  class the group settles on rises while the others dip, so the correct
  trajectory escapes the convex hull/space formed by the group's initial beliefs.
Classical consensus cannot model this, as every opinion at every round stays
  inside that initial hull.
We hypothesize a hidden driving force these classical models omit, and argue it
  is the LLM's own internal belief.

\textbf{LLM deliberation is a closed-loop system:}
Multi-agent systems are commonly cast as either open-loop or
  closed-loop~\cite{TODO-astrom}.
An open-loop system evolves from its own state and inputs with no external control fed 
  to steer it, whereas a closed-loop system adds a feedback/control term that drives the
  state toward a reference, letting the trajectory settle away from the 
  average of its inputs.
LLM deliberation's implementation, in prior work cited above and in this work, uses no such feedback. Each agent
  only exchanges opinions with its neighbours, so the update is a function of
  the agents' states alone. 
This is exactly why we apply classical open loop  
  consensus dynamics, but find that it cannot reproduce the escape above.
We therefore model deliberation as a closed-loop system in which each agent is
  assigned a hidden anchor, its internal belief. This anchor is the control
  signal absent from the open-loop view, and it is what lets the 
  trajectory, leave the initial hull.

% We show the anchor is identifiable from observed trajectories alone.
% Validating the fit on held-out runs turns the procedure into a model selection
%   test that detects when a hidden anchor governs deliberation.
% Across three open-weight families the effect is a spectrum.
% The mechanism is uniform, every family anchors with comparable strength, but
%   the families differ in where the anchor sits, and only a distant anchor lets
%   deliberation escape the hull.

\textbf{Contributions:}
Our contributions are the following.
(a) A new closed-loop interaction dynamics for multi-agent LLM deliberation
  that includes a hidden per-agent anchor.
(b) An empirical characterisation of where this dynamics converges, showing that
  deliberation settles within the convex hull defined by the agents' recovered
  anchor beliefs (\S\ref{sec:geometry}, Figure~\ref{fig:hull}).
(c) A system identification and held-out validation procedure that recovers the
  anchor from trajectories and acts as a model selection test, showing that
  anchor strength is a spectrum across model families rather than a uniform
  property.
  % with the linear baselines selected exactly where no distant anchor
  %is present.

\section{Related Work}

\subsection{Multi-Agent LLM Debate}
\label{MAS-Debate}
Multi-agent deliberation among LLMs improves reasoning and accuracy.
\citet{TODO-du-debate} let model instances debate over multiple rounds,
  reporting gains on mathematical and strategic reasoning.
\citet{TODO-liang} push agents into adversarial ``tit-for-tat'' exchanges
  refereed by a judge, countering degeneration of thoughts.
ChatEval~\cite{TODO-chateval} turns role-based multi-agent debate into a
  stronger automatic evaluator.
% Beyond accuracy, \citet{TODO-khan} show that optimizing debaters for
%   persuasiveness makes weaker judges more truthful, positioning debate as a
%   scalable-oversight mechanism.
In these works the round-by-round trajectory of belief is never
  modelled, and there is no account of why deliberation converges
  where it does. Closing that gap is the contribution of this paper.

\subsection{Opinion Dynamics and Consensus}
\label{sec:opinion}
The trajectory of interacting beliefs is formalised in the opinion-dynamics
  literature.
DeGroot learning~\cite{TODO-degroot} replaces each opinion with a weighted
  average of its neighbours', the Friedkin--Johnsen model~\cite{TODO-fj}
  adds per-agent anchoring to the initial opinion, and the
  Hegselmann--Krause~\cite{TODO-hk} bounded-confidence rule averages only
  over sufficiently close neighbours.
All three share a convex-hull bound. Every update is a convex combination
  of current (and, for Friedkin--Johnsen, initial) opinions, so no
  coordinate can leave the convex hull of the initial opinions.
We observe LLM deliberation that violates this bound for some model families,
  which a classical linear rule cannot reproduce and which motivates the
  closed-loop model we develop.

\subsection{Opinion-Dynamics Simulation via LLMs}
A separate line of work uses LLM agents to simulate classical
  opinion dynamics and asks whether LLMs reproduce human social behaviour.
OpinioNet~\cite{TODO-opinionet} models ideological community agents
  updating through external-event influence, network structure, and opinion
  inertia, outperforming Friedkin--Johnsen, Hegselmann--Krause, and
  Deffuant--Weisbuch on real social-media trajectories.
\citet{TODO-HeAtAl} run multi-round LLM dialogues that retain each
  agent's initial opinion and conclude that LLM opinion formation is
  ``largely consistent with Friedkin--Johnsen''.
\citet{TODO-chuang} report that networked LLM agents bias toward accurate
  consensus and fragment only when prompted with confirmation bias.
All three impose a prescribed classical rule and use the LLM for
  simulation. None analyse the deliberation system itself or recover a
  per-agent latent state from trajectories.
%TODO: cross-reference these works again in the Results section.
The contrast is sharpest against \citet{TODO-HeAtAl}: their Friedkin--Johnsen
  consistency predicts every coordinate stays inside the initial hull, yet we
  find this holds only for families whose recovered anchor coincides with the
  initial opinion and fails for those whose anchor lies elsewhere, where the
  gold-class coordinate leaves the hull.

% \subsection{Sycophancy and Persuasion in LLMs}
% Sharma et al. sycophancy. Connect: our per-agent anchor strength beta_i is
% exactly the consensus-pull vs internal-prior trade-off that sycophancy
% measures. Sets up the (optional) beta-vs-sycophancy correlation.

\section{Problem Setting}
\label{sec:problem}
% Formal notation: n agents on graph G, round k, probability vector
% x_i(k) in simplex over d classes, recovered from token logits. Define the
% deliberation as a discrete-time process. Keep it tight -- this is setup, not
% results.

We study a population of $n$ LLM agents that deliberate over a
  fixed multiple-choice question with $d$ possible answer classes.
The agents communicate over a directed graph $G=(V,E)$ with
  $V=\{1,\dots,n\}$. We write $\mathcal{N}_i=\{j:(j,i)\in E\}$ for the
  neighbours whose outputs agent $i$ observes, and $A\in\{0,1\}^{n\times n}$
  for the adjacency matrix with $A_{ij}=1$ iff $j\in\mathcal{N}_i$.
Deliberation proceeds in synchronous rounds $k=0,1,\dots,K$.
At each round, agent $i$'s belief is a probability vector
\begin{equation}
  \mathbf{x}_i(k)=\big(x_{i,1}(k),\dots,x_{i,d}(k)\big)\in\Delta^{d-1},
\end{equation}
  where $\Delta^{d-1}=\{\mathbf{p}\in\mathbb{R}^d_{\ge0}:\sum_{c}p_c=1\}$ is
  the probability simplex over the $d$ classes.
We let $g\in\{1,\dots,d\}$ denote the index of the gold (correct) class and
  refer to $x_{i,g}(k)$ as the gold-class coordinate.

\subsection{Multi-Agent Deliberation Protocol}
\label{sec:protocol}
% Round-robin mechanics, ring topology, prompt structure (own + neighbour
% previous outputs), how a probability vector is extracted per round.

\begin{figure*}[t]
    \centering
    \includegraphics[width=\textwidth]{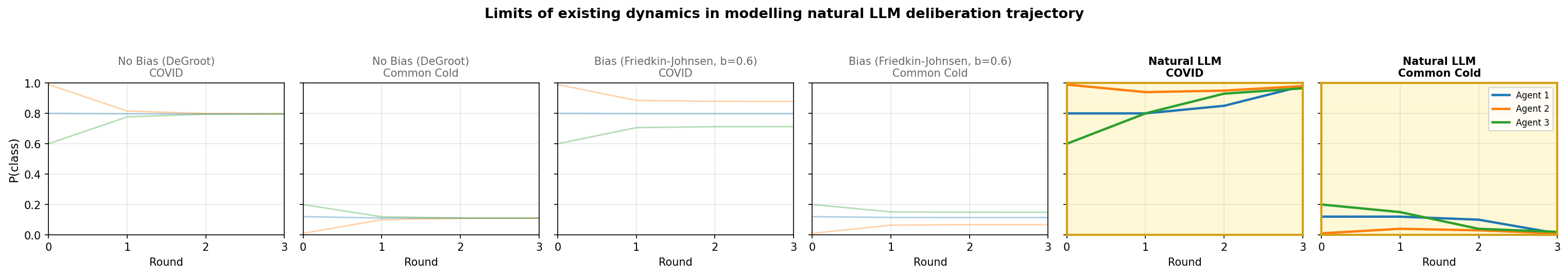}
    \caption{Probability trajectories (snippet) for the gold class (COVID) and a
      competing class (Common Cold) under the protocol of
      \S\ref{sec:protocol} ($n{=}3$ agents, ring topology). Left and middle
      panels show the open-loop baselines (DeGroot and Friedkin--Johnsen)
      initialised with the agents' real first-round beliefs: every class stays
      inside the band of its initial values, as Property~\ref{prop:hull}
      guarantees. Right panels (highlighted) show the real LLM round-robin
      deliberation: the gold-class probability grows past the maximum
      initial value across all agents, the empirical anomaly
      \eqref{eq:escape} that the rest of the paper explains.}
    \label{fig:dynamics_limits}
\end{figure*}

We instantiate $G$ as a directed ring where agent $i$ observes its
  ring-predecessor, $\mathcal{N}_i=\{(i-1)\bmod n\}$, so every agent has
  exactly one neighbour and influence propagates around the cycle.
At round $0$ each agent answers the question independently, producing its
  initial belief $\mathbf{x}_i(0)$.
At every subsequent round $k>0$ agent $i$ is re-prompted with (i) the original
  question, (ii) its own previous answer, and (iii) the previous-round answer
  of its neighbour in $\mathcal{N}_i$, and is asked to reconsider and re-rank
  the candidate classes.
The model returns a ranked top-$5$ list with self-reported probabilities. We
  map each entry to the default class label and renormalise.
This yields, for every run, a fully observed discrete-time trajectory which is the object the rest of the
  paper models. Figure~\ref{fig:dynamics_limits} (right) shows one such
  trajectory.

\subsection{Open-Loop Consensus Baselines}
\label{sec:openloop}
% Define DeGroot and Friedkin-Johnsen updates. State and prove (or cite) the
% convex-hull bound Property. Conclude: structurally cannot reproduce the
% anomaly -- this is impossibility, not a weak baseline.

Under DeGroot dynamics~\cite{TODO-degroot} each agent moves a fixed
  step $\varepsilon\in(0,1)$ towards the average of its neighbours.
\begin{equation}
  \begin{split}
    \mathbf{x}_i(k{+}1)=\;&\mathbf{x}_i(k)\\
      &+\varepsilon\!\!\sum_{j\in\mathcal{N}_i}\!\!
        A_{ij}\big(\mathbf{x}_j(k)-\mathbf{x}_i(k)\big),
  \end{split}
  \label{eq:degroot}
\end{equation}
  % which is a row-stochastic convex update of the current opinions.
The Friedkin--Johnsen model~\cite{TODO-fj} adds per-agent
  stubbornness/bias: with susceptibility $\lambda\in[0,1]$,
\begin{equation}
  \begin{split}
    \mathbf{x}_i(k{+}1)=\;&\lambda\Big(\mathbf{x}_i(k)+\varepsilon\!\!
      \sum_{j\in\mathcal{N}_i}\!\!A_{ij}\big(\mathbf{x}_j(k)-\mathbf{x}_i(k)\big)\Big)\\
      &+(1-\lambda)\,\mathbf{x}_i(0),
  \end{split}
  \label{eq:fj}
\end{equation}
  so each agent is pulled back towards its own initial opinion
  $\mathbf{x}_i(0)$.
% Both rules are open-loop: the update depends only on observed opinions
%   and never on a latent agent-specific target.
% They share a structural invariant.

\begin{property}[Convex-hull bound]
\label{prop:hull}
For the updates in \eqref{eq:degroot} and \eqref{eq:fj} with
  $\varepsilon$ small enough that every update is a convex combination, and
  for every class $c$,
\begin{equation}
  \min_{1\le j\le n} x_{j,c}(0)\;\le\; x_{i,c}(k)\;\le\;\max_{1\le j\le n} x_{j,c}(0)
  \quad\forall i,\,\forall k.
\end{equation}
That is, no coordinate of any opinion can ever leave the convex hull of the
  initial opinions $\mathrm{conv}\{\mathbf{x}_j(0)\}_j$.
\end{property}

Figure~\ref{fig:dynamics_limits} (left and middle) illustrates this
  initialised with the agents' real first-round beliefs, both updates keep every
  class inside the band of its initial values. These baselines therefore do not
  merely fit poorly. They are \emph{structurally incapable} of any trajectory
  that leaves the initial hull.

\paragraph{Empirical anomaly: escape from the convex hull.}
The motivating observation of this paper is that real deliberation violates
  Property~\ref{prop:hull} on the gold-class coordinate. Across runs we
  repeatedly observe
\begin{equation}
  \max_{i,\,k}\; x_{i,g}(k)\;>\;\max_{j}\; x_{j,g}(0),
  \label{eq:escape}
\end{equation}
  the gold-class probability rising strictly above the largest value any agent
  held initially, so the trajectory escapes $\mathrm{conv}\{\mathbf{x}_j(0)\}_j$.
The fraction of runs in which this happens depends strongly on the model
  family (\S\ref{sec:escape}). A hidden, per-agent driving force must therefore
  be added to the dynamics.

\section{Hidden-Anchor Model}
\label{sec:anchor}
% Core model section. Build from the baseline that fails to the closed-loop
% model that works (steady-state proof moved to empirical \S\ref{sec:geometry}).

The missing ingredient behind the escape~\eqref{eq:escape} is a force that does
  not depend on the observed opinions at all. We therefore augment the
  consensus update with a hidden, per-agent anchor $\mathbf{b}_i$, the agent's
  own internal belief, that continually pulls $\mathbf{x}_i$ towards
  $\mathbf{b}_i$ regardless of its neighbours.
% NOTE: this is not the case as it depends on the llm family. need to change this.
We find empirically that this closed-loop system settles into
  $\mathrm{conv}\{\mathbf{b}_i\}_i$ rather than
  $\mathrm{conv}\{\mathbf{x}_j(0)\}_j$ (\S\ref{sec:geometry}), reproducing the
  anomaly that linear consensus forbids.

% We call this system closed-loop because in the control-theoretic sense the
%   update is driven by feedback. Two feedback
%   paths are present. First, at the protocol level the agent's own output is
%   re-injected as its input every round (\S\ref{sec:protocol}). Second, the anchor term
%   $-\beta_i(\mathbf{x}_i-\mathbf{b}_i)$ is proportional error feedback that
%   regulates the opinion towards a latent internal reference $\mathbf{b}_i$.

\subsection{Hidden-Anchor Update Rule}
% The boxed closed-loop equation: consensus pull (scaled DeGroot) + per-agent
% anchor pull beta_i(x_i - b_i). Explain each term's LLM interpretation
% (neighbour prompt vs internal reasoning prior).

Each agent updates as
\begin{equation}
  \begin{split}
    \mathbf{x}_i(k{+}1)=\;&\mathbf{x}_i(k)\\
      &-\alpha\!\!\sum_{j\in\mathcal{N}_i}\!\!
        A_{ij}\big(\mathbf{x}_i(k)-\mathbf{x}_j(k)\big)\\
      &-\beta_i\big(\mathbf{x}_i(k)-\mathbf{b}_i\big),
  \end{split}
  \label{eq:anchor}
\end{equation}
with a single shared consensus gain $\alpha\ge0$, a per-agent anchor gain
  $\beta_i\ge0$, and a hidden anchor $\mathbf{b}_i\in\Delta^{d-1}$.
The first correction is the consensus pull: a scaled DeGroot step
  ($\alpha$ plays the role of $\varepsilon$ in Eq.~\eqref{eq:degroot}) that
  models the agent reacting to the neighbour opinion injected into its prompt.
The second is the anchor pull: a persistent attraction towards the
  agent's own latent reasoning prior $\mathbf{b}_i$ that the prompt context
  does not override and that never appears in the observed opinions.
This prior is not arbitrary: an LLM carries inherent beliefs fixed by its
  pre-training data and architecture, much as a person reasons from the
  background knowledge and predispositions they bring to a
  discussion~\cite{TODO-anchoring}.

Equation~\eqref{eq:anchor} strictly generalises both baselines: setting
  $\beta_i=0$ recovers DeGroot~\eqref{eq:degroot}, and replacing the latent
  $\mathbf{b}_i$ by the observed initial opinion $\mathbf{x}_i(0)$
  recovers Friedkin--Johnsen~\eqref{eq:fj}.

\section{System Identification}
\label{sec:sysid}

The hidden-anchor model~\eqref{eq:anchor} is parameterised by
  $(\alpha,\beta_i,\mathbf{b}_i)$. This section recovers those
  parameters from observed trajectories.

\textbf{Linear reparameterisation.}
Let $\Delta\mathbf{x}_i(k):=\mathbf{x}_i(k{+}1)-\mathbf{x}_i(k)$ and
  $\boldsymbol{\gamma}_i:=\beta_i\mathbf{b}_i$.
With this reparameterisation, \eqref{eq:anchor} becomes
\begin{equation}
  \Delta\mathbf{x}_i(k) = -\alpha\!\!\sum_{j\in\mathcal{N}_i}\!\!
    A_{ij}\big(\mathbf{x}_i(k)-\mathbf{x}_j(k)\big)
    -\beta_i\mathbf{x}_i(k)+\boldsymbol{\gamma}_i,
  \label{eq:anchor-linear}
\end{equation}
  linear in $\theta=(\alpha,\{\beta_i\},\{\boldsymbol{\gamma}_i\})$.
Stacking \eqref{eq:anchor-linear} across all agents, rounds, class
  coordinates, and runs yields an overdetermined system
  $\mathbf{A}\theta=\mathbf{y}$, solved by ordinary least-squares.
Fit quality is reported as
\begin{equation}
  R^2(\Delta\mathbf{x})=1-
    \frac{\sum\|\Delta\mathbf{x}_i(k)-\widehat{\Delta\mathbf{x}}_i(k)\|^2}
         {\sum\|\Delta\mathbf{x}_i(k)-\overline{\Delta\mathbf{x}}\|^2},
  \label{eq:r2}
\end{equation}
  evaluated on one-step displacements so that the metric is not dominated by
  the trivial $\mathbf{x}_i(k{+}1)\approx\mathbf{x}_i(k)$ baseline.

\textbf{Identifiability.}
Equation~\eqref{eq:anchor-linear} is linear in
  $(\alpha,\beta_i,\boldsymbol{\gamma}_i)$ but not in
  $(\alpha,\beta_i,\mathbf{b}_i)$, because the latter pair enters only through
  the product $\boldsymbol{\gamma}_i=\beta_i\mathbf{b}_i$.

% \begin{proposition}[Identifiability]
% \label{prop:identifiability}
% $(\alpha,\{\beta_i\},\{\mathbf{b}_i\})$ is uniquely recoverable from
%   observed trajectories if and only if (i) the design matrix $\mathbf{A}$ has
%   full column rank, and (ii) either $\mathbf{b}_i$ is known a priori for some
%   agent, or at least two runs with distinct anchor conditions are observed.
%   From a single run with no prior on $\mathbf{b}_i$, only the products
%   $\boldsymbol{\gamma}_i$ and $\alpha$ are identifiable.
% \end{proposition}

% We satisfy (i) by stacking across classes and runs (verified numerically per
%   cell) and (ii) by running each $(\text{topic},\text{model})$ cell with
%   multiple seeds inducing distinct $\mathbf{x}_i(0)$.

\textbf{Anchor recovery.}
Anchors are recovered as $\tilde{\mathbf{b}}_i=\hat{\boldsymbol{\gamma}}_i/
  \hat\beta_i$ and projected onto the simplex,
  $\hat{\mathbf{b}}_i=\Pi_{\Delta^{d-1}}(\tilde{\mathbf{b}}_i)$, by the
  $O(d\log d)$ algorithm of \citet{TODO-duchi}.
When $\hat\beta_i$ is small the division is ill-conditioned. We report
  $\hat\beta_i$ alongside every recovered anchor and flag agents below a
  threshold as unreliable.

\section{Experimental Setup}
\label{sec:setup}

We run all experiments on three open-weight instruction-tuned LLMs,
  Llama-3.1-70B-Instruct~\cite{TODO-llama3}, Qwen3-32B~\cite{TODO-qwen3},
  and gpt-oss-20b~\cite{TODO-gptoss}, on a symptom$\rightarrow$disease
  diagnosis task, in which each agent ranks candidate diseases for a
  symptom set drawn from a $42$-class diagnosis benchmark~\cite{TODO-symptom}.
We use $10$ cases, each a distinct target disease, to span a range of
  initial-opinion geometries.
Each $(\text{model},\text{case})$ cell uses the protocol of
  \S\ref{sec:protocol} with $n{=}3$ agents on a directed ring, $K{=}5$
  rounds, and $3$ random seeds, yielding $90$ deliberation trajectories
  ($30$ per model).
We implement deliberation as a round-robin message-passing graph in
  LangGraph~\cite{langgraph}, where
  each agent is a node that reads its own and its neighbour's previous response
  and emits an updated prediction.
The interaction protocol follows the multi-agent debate paradigm of
  \citet{TODO-du-debate}, and most closely resembles the more recent
  deliberation frameworks built on the consensus protocol
  of~\citet{apurba2025}.
Evaluation metrics ($R^2(\Delta\mathbf{x})$, held-out MSE, bootstrap CIs,
  hull-containment rate), full decoding
  hyperparameters, prompt templates, and checkpoint identifiers are deferred
  to Appendix~\ref{app:setup}.
% TODO(camera-ready): de-anonymise. Anonymous repo link for double-blind review.
Code, data, and analysis scripts are available at the provided zip file. 

\subsection{Analysis Procedures}
\label{sec:procedures}
Every stored trajectory is post-processed by a fixed battery of analyses,
  referenced by: Experiment + letter throughout the paper:
\begin{description}[leftmargin=1.6em,itemsep=2pt,topsep=2pt]
  \item[\textbf{A. Open-loop baseline.}] Initialise the linear consensus rules
    (\S\ref{sec:openloop}) with the agents' first-round beliefs and simulate
    forward, confirming they stay inside $\mathrm{conv}\{\mathbf{x}_j(0)\}$
    (Property~\ref{prop:hull}).
  \item[\textbf{B. Observed vs.\ linear.}] Compare the real LLM trajectory
    against the round-by-round linear-consensus prediction, exposing where the
    open-loop model fails.
  \item[\textbf{C. System identification.}] Fit the hidden-anchor update
    (\S\ref{sec:sysid}) by least squares, recovering
    $(\hat\alpha,\hat\beta_i,\hat{\mathbf{b}}_i)$ and the in-sample
    $R^2(\Delta\mathbf{x})$.
  \item[\textbf{D. Anchor drift.}] Refit on an early and a late window of
    rounds and compare anchors, distinguishing compliance (stable anchor) from
    internalisation (moving anchor).
  \item[\textbf{E. Bootstrap CIs.}] Block-bootstrap the Experiment~C fit per
    run for parameter confidence intervals (\S\ref{sec:uncertainty}).
  \item[\textbf{F. Held-out cross-run validation.}] Fit on a subset of a
    problem's seeds and predict a held-out seed (leave-one-out ensemble),
    measuring whether recovered parameters  {generalise}.
\end{description}
We additionally fit  {nested} restrictions of the anchor model with the
  same least-squares machinery as Experiment~C, DeGroot ($\beta_i{=}0$) and
  Friedkin--Johnsen ($\mathbf{b}_i{=}\mathbf{x}_i(0)$), so that the three
  models are compared on an identical target both in-sample and under the
  held-out protocol of Experiment~F. This nested comparison is a structural
  ablation of the anchor mechanism.

\section{Analysis}
\label{sec:results}
% % Follow the paper's narrative arc: open-loop fails -> closed-loop fits ->
% % parameters validate -> geometry confirms. Each subsection =
% % one claim + the figure/number that supports it.

We analyse the symptom$\rightarrow$disease deliberation benchmark: three
  open-weight models (Llama-3.1-70B, Qwen3-32B, gpt-oss-20b), $10$ distinct
  target diseases, and $3$ random seeds per problem, for $30$ independent
  deliberation runs per model ($90$ total; $n{=}3$ agents, $K{=}5$ reflection
  rounds, ring topology). Cross-run quantities use the leave-one-seed-out
  protocol of Experiment~F.

\subsection{Trajectories Are Not Uniform Across Model Families}
\label{sec:traj-obs}
The motivating observation comes from inspecting the trajectories directly
  (Experiments~A and~B; Figure~\ref{fig:dynamics_limits}). The open-loop
  baselines, initialised with real first-round beliefs, never leave the band of
  initial opinions, as Property~\ref{prop:hull} guarantees. The  {real}
  deliberation behaves very differently, but not uniformly across models.
  Llama-3.1-70B produces sharp, non-monotone swings in the gold-class
  probability that overshoot the initial band substantially. Qwen3-32B and
  gpt-oss-20b are markedly flatter: their gold-class coordinate moves little
  and is, in many runs, well described by a linear convex update.
  Quantitatively, taking the per-run range to be the spread (maximum minus
  minimum) of the gold-class probability across rounds $0$--$K$ within a single
  run, its mean over the $30$ runs is $0.26$ for Llama versus $0.09$ (Qwen) and
  $0.12$ (gpt-oss) (Table~\ref{tab:app-range}), and the total variation follows
  the same ordering. This rules out a single universal claim and
  motivates a  {per-family} comparison of the linear baselines against the
  hidden-anchor model, in-sample and, crucially, under held-out
  validation.

\begin{table*}[t]
\centering
\small
\begin{tabular}{lcccccccc}
\toprule
& \multicolumn{3}{c}{In-sample $R^2(\Delta\mathbf{x})$ (mean)}
& \multicolumn{3}{c}{Held-out $R^2(\Delta\mathbf{x})$ (mean)} & \\
\cmidrule(lr){2-4}\cmidrule(lr){5-7}
Model & DeGroot & FJ & Anchor & DeGroot & FJ & Anchor & Anchor sel.\ \% \\
\midrule
Llama-3.1-70B & $0.12$ & $0.30$ & $\mathbf{0.86}$ & $0.05$ & $0.04$ & $\mathbf{0.44}$ & $80$ \\
Qwen3-32B     & $0.26$ & $0.32$ & $\mathbf{0.64}$ & $\mathbf{0.10}$ & $0.10$ & $0.08$ & $70$ \\
gpt-oss-20b   & $0.30$ & $0.37$ & $\mathbf{0.61}$ & $\mathbf{0.13}$ & $0.12$ & $-0.94$ & $20$ \\
\bottomrule
\end{tabular}
\caption{Nested model comparison on observed deliberation trajectories
  ($30$ runs per model). The three models are the hidden-anchor update and its
  restrictions DeGroot ($\beta_i{=}0$) and Friedkin--Johnsen
  ($\mathbf{b}_i{=}\mathbf{x}_i(0)$), all fit by the same least squares
  (Experiment~C). Both blocks report the  {mean} $R^2(\Delta\mathbf{x})$:
  in-sample over the $30$ runs per model, held-out over the $10$ disease groups.
  {Held-out} $R^2$ is the leave-one-seed-out reconstruction of the step changes
  of an unseen seed (Experiment~F). A negative held-out value means the fitted
  model predicts the step changes worse than their own mean, so the extra
  freedom does more harm than good (gpt-oss, see text).
  ``Anchor sel.\ \%'' is the fraction of the $10$ disease groups in which the
  full anchor model attains the best held-out $R^2$. Best per block in bold.}
\label{tab:nested}
\end{table*}

\subsection{In-Sample Fit Favours the Anchor Model, but Cannot Settle It}
\label{sec:insample}
Fit in-sample, the full hidden-anchor model dominates both baselines for every
  family (Table~\ref{tab:nested}, left block): $R^2(\Delta\mathbf{x})$ rises
  from $0.12$--$0.30$ (DeGroot) and $0.30$--$0.37$ (Friedkin--Johnsen) to
  $0.61$--$0.86$ (full). DeGroot is insufficient everywhere, confirming that
  pure neighbour-averaging does not describe deliberation. But the full model
  also carries the most free parameters (the $n$ latent anchors), so an
  in-sample advantage is expected by construction and  {cannot} on its own
  establish that the anchors are real rather than fit to noise. The decisive
  test is whether the recovered parameters generalise.

\subsection{Held-Out Validation Reveals Fit}
\label{sec:heldout}
Under leave-one-seed-out validation (Table~\ref{tab:nested}, right block) the
  three families separate sharply. Full breakdown in Table~\ref{tab:app-selection}.
\begin{itemize}[leftmargin=1.4em,itemsep=1pt,topsep=2pt]
  \item \textbf{Llama-3.1-70B:} the full model's mean held-out
    $R^2$ is $0.44$ against $\approx0.05$ for both baselines, and it is selected
    in $8/10$ groups. The latent anchor is a genuine, transferable property:
    parameters fit on two seeds predict an unseen seed where linear consensus
    cannot.
  \item \textbf{Qwen3-32B:} the full model is still selected in $7/10$ groups,
    but its mean held-out $R^2$ ($0.08$) sits at or just below the baselines
    ($\approx0.10$): a weak anchor signal on near-linear dynamics, where the few
    overfit folds already pull the mean down. The transfer here lives in the
    selection count rather than in the average fit.
  \item \textbf{gpt-oss-20b:} the baselines win (mean $\approx0.13$ vs.\
    $-0.94$, full selected in only $2/10$ groups). The full model's mean
    held-out $R^2$ is sharply  {negative}: on an unseen seed it predicts the
    step changes worse than simply using their mean, so adding the hidden anchor
    does more harm than good. Its deliberation is essentially linear consensus,
    with no transferable latent anchor.
\end{itemize}
We therefore read held-out validation as a  \textbf{model-selection} criterion:
  it determines  {when} a hidden anchor governs deliberation rather than
  imposing one. That the procedure selects the linear baselines for gpt-oss,
  and penalises the over-parameterised anchor model exactly where no anchor is
  present, is evidence that the recovered anchors elsewhere are not an
  artefact of model capacity.

\begin{table}[t]
\centering
\small
\begin{tabular}{lcccc}
\toprule
Model & $\bar{\hat\beta}$ & $\hat{\mathbf{b}}$ margin & \% out & SS cont.\ \% \\
\midrule
Llama-3.1-70B & $0.34$ & $\mathbf{0.33}$ & $\mathbf{92}$ & $74$ \\
Qwen3-32B     & $0.36$ & $0.10$ & $37$ & $74$ \\
gpt-oss-20b   & $0.34$ & $0.10$ & $48$ & $60$ \\
\bottomrule
\end{tabular}
\caption{Recovered-anchor geometry (Experiment~C, $30$ runs per model).
  $\bar{\hat\beta}$ is the mean per-agent anchor gain. ``$\hat{\mathbf{b}}$
  margin'' is the median overshoot of the recovered anchors past the initial
  band $\mathrm{conv}\{\mathbf{x}_j(0)\}$ on the gold coordinate. ``\% out'' is
  the fraction of runs whose anchors lie more than $0.10$ outside that band.
  ``SS cont.''\ is the steady-state (final-round) containment rate in
  $\mathrm{conv}\{\hat{\mathbf{b}}_j\}$ at tolerance $0.05$.}
\label{tab:geometry}
\end{table}

\subsection{Recovered Anchor Geometry Explains the Gradient}
\label{sec:geometry}
Why does the same procedure certify an anchor for Llama but reduce to a linear
  baseline for gpt-oss? The recovered parameters answer this directly
  (Table~\ref{tab:geometry}). The anchor  {gain} does not discriminate:
  the mean per-agent gain $\bar{\hat\beta}$ is essentially flat across families
  ($0.34$--$0.36$). What differs is the anchor  {location}. For Llama the
  recovered anchors $\hat{\mathbf{b}}_i$ lie far outside the band of initial
  opinions on the gold coordinate (median margin $0.33$, outside in $92\%$ of
  runs). For Qwen and gpt-oss they sit essentially  {at} the initial band
  (median $0.10$). An anchor that coincides with the initial opinion,
  $\hat{\mathbf{b}}_i\!\approx\!\mathbf{x}_i(0)$, is exactly the
  Friedkin--Johnsen special case of \eqref{eq:anchor} (cf.\ \eqref{eq:fj}),
  which explains why the held-out criterion selects FJ for gpt-oss: that model
   {is} anchored, to its own starting belief, and so cannot leave the
  initial hull.

\begin{figure}[t]
\centering
\begin{subfigure}{\linewidth}
  \centering
  \includegraphics[width=\linewidth]{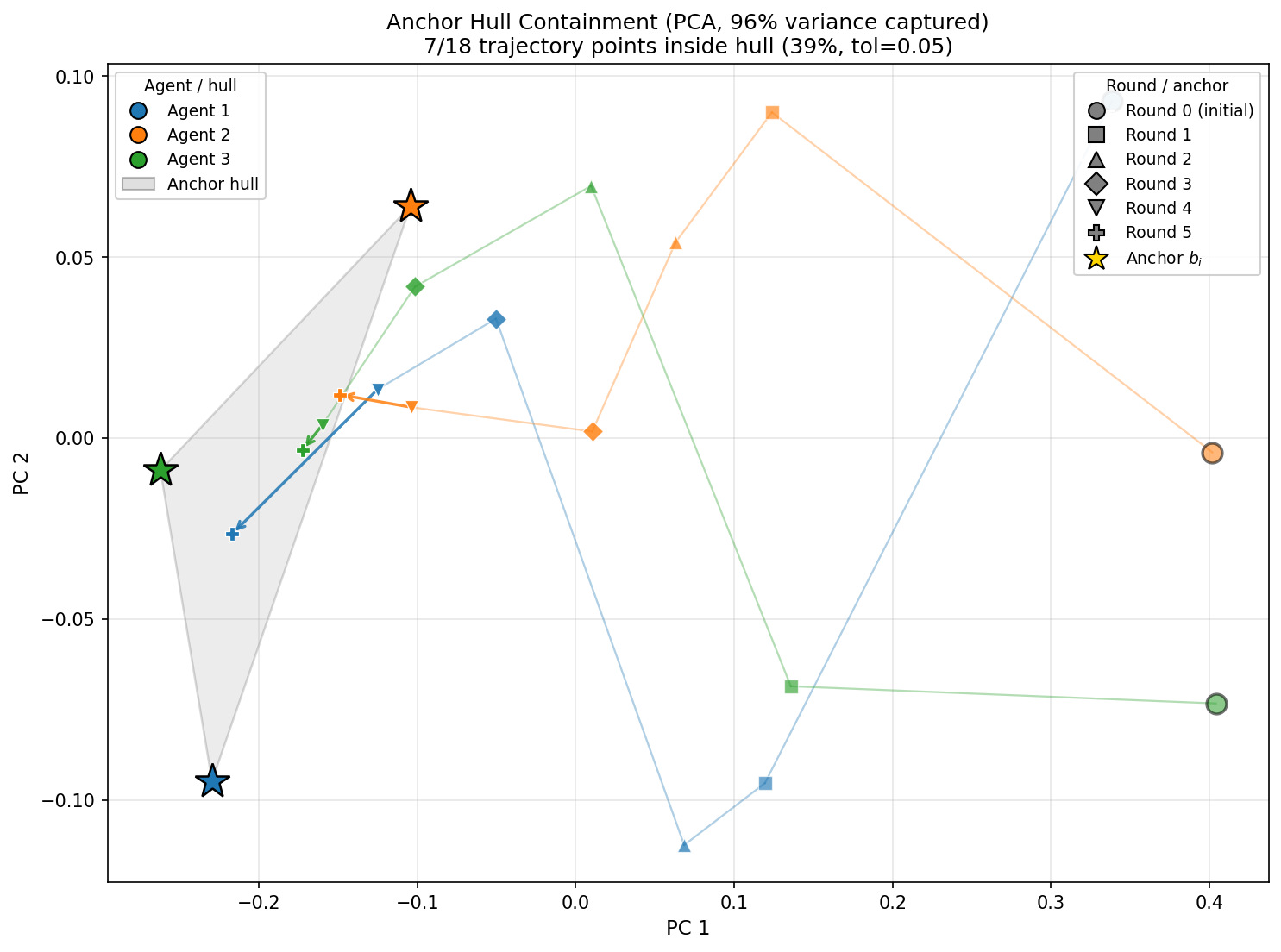}
  \caption{Llama-3.1-70B: opinions settle inside $\mathrm{conv}\{\hat{b}_i\}$.}
  \label{fig:hull-llama}
\end{subfigure}\\[2pt]
\begin{subfigure}{\linewidth}
  \centering
  \includegraphics[width=\linewidth]{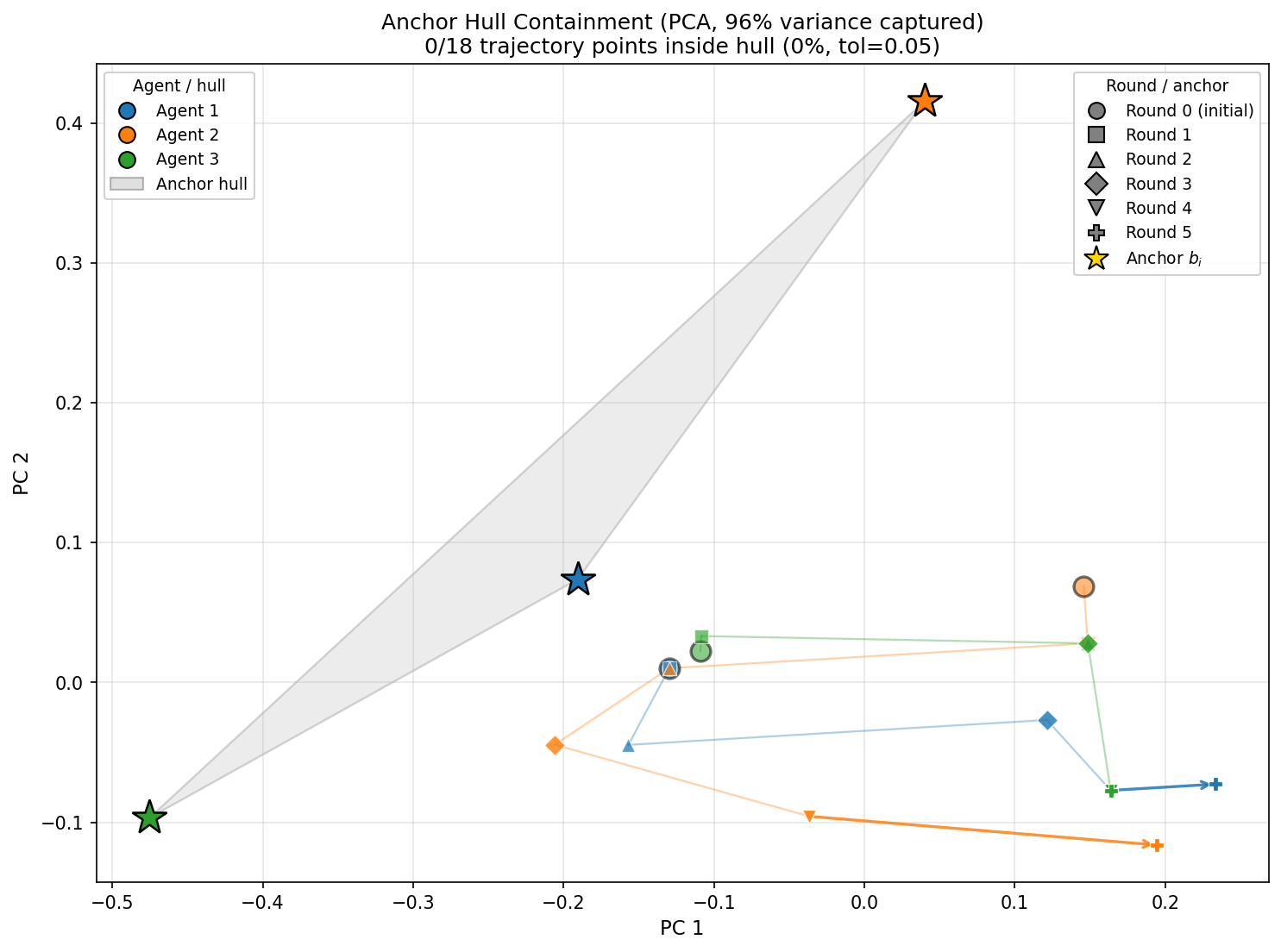}
  \caption{Qwen3-32B: opinions never enter the anchor hull ($0/18$).}
  \label{fig:hull-qwen}
\end{subfigure}
\caption{Recovered anchor hull $\mathrm{conv}\{\hat{b}_i\}$ (shaded) and the
  deliberation trajectory (PCA-projected; rounds $0$--$5$) for one case
  (GERD, seed~2). Llama's later-round opinions converge into the hull
  (steady state contained), whereas Qwen's stay outside throughout. The gpt-oss
  case matches Qwen (omitted, $0/18$ inside, $75\%$ variance captured). This is the
  geometric counterpart of the steady-state containment rates in
  Table~\ref{tab:geometry}.}
\label{fig:hull}
\end{figure}

This geometry also fixes where deliberation converges. Empirically the
  final-round opinions settle inside the recovered anchor hull
  $\mathrm{conv}\{\hat{\mathbf{b}}_j\}$ rather than the initial one, in $74\%$
  (Llama), $74\%$ (Qwen), and $60\%$ (gpt-oss) of runs, far above the
  $29$--$48\%$ whole-trajectory rate, since intermediate iterates overshoot
  before settling. The trajectory therefore escapes
  $\mathrm{conv}\{\mathbf{x}_j(0)\}$ precisely when some recovered anchor lies
  outside it. This containment is the empirical content of contribution~(b).
  Figure~\ref{fig:hull} shows a representative case (GERD, seed~2): Llama's
  opinions converge into the anchor hull while Qwen's never enter it. The
  gpt-oss case is indistinguishable from Qwen on this run (no point inside the
  hull, $75\%$ PCA variance captured). 
  % Finally, this resolves why the in-sample $\hat\beta$
  % could not predict held-out success (\S\ref{sec:heldout}): anchor strength is
  % uniform across families, so only the recovered anchor  {position} carries
  % the signal.

\subsection{Hull Escape Is Family-Dependent}
\label{sec:escape}
The geometric anomaly of \eqref{eq:escape}, the gold-class probability
  leaving $\mathrm{conv}\{\mathbf{x}_j(0)\}$, which Property~\ref{prop:hull}
  forbids for the linear baselines, tracks the same ordering
  (Table~\ref{tab:escape}). We call a run an escape at tolerance $\tau$ when its
  escape margin, the largest overshoot of any agent's gold-class probability
  past the round-$0$ band $\mathrm{conv}\{\mathbf{x}_j(0)\}$ in any later round
  (Table~\ref{tab:escape}), exceeds $\tau$, with $\tau$ measured in probability
  units. The verdict depends on the tolerance required: at a negligible
  $\tau{=}0.02$ all three models escape in $\approx75\%$ of runs, but at a
  meaningful $\tau{=}0.10$ only
  Llama escapes substantially ($77\%$ of runs; mean overshoot $0.22$), while
  Qwen and gpt-oss escape in $\approx25\%$ of runs with mean overshoot
  $\approx0.07$. The hull-violating dynamics are thus concentrated in the family for
  which held-out validation certifies a latent anchor, and follow
  directly from its anchor geometry (\S\ref{sec:geometry}). Only Llama's
  recovered anchors lie far enough outside the initial hull to drive the
  trajectory past it.

\begin{table}[t]
\centering
\small
\begin{tabular}{lccc}
\toprule
Model & Mean & Median & \% runs $>0.10$ \\
\midrule
Llama-3.1-70B & $0.22$ & $0.20$ & $77$ \\
Qwen3-32B     & $0.08$ & $0.06$ & $23$ \\
gpt-oss-20b   & $0.07$ & $0.06$ & $27$ \\
\bottomrule
\end{tabular}
\caption{Gold-coordinate escape margin beyond
  $\mathrm{conv}\{\mathbf{x}_j(0)\}$ (the band of initial opinions), over $30$
  runs per model. The margin is the largest overshoot of any agent's
  gold-class probability past the round-$0$ band in any later round.}
\label{tab:escape}
\end{table}

\subsection{Anchor Behaviour is a Spectrum}
\label{sec:gradient}
  Hidden-anchor behaviour is not all-or-nothing but a  {spectrum across
  model families}: Llama-3.1-70B $\gg$ Qwen3-32B $>$ gpt-oss-20b, with gpt-oss
  at the linear-consensus (Friedkin--Johnsen) end of the scale. The mechanism
  is uniform, every family anchors to a latent prior with comparable
  strength, but only when that prior sits far from the initial opinions
  does deliberation escape the convex hull and demand the full
  closed-loop model. Our held-out criterion recovers that boundary without
  supervision.

\begin{table}[t]
\centering
\small
\begin{tabular}{lcccc}
\toprule
Model & $\bar{\hat\beta}\!\le\!0$ & $\hat\beta$ CI $\ni 0$ & $\hat\alpha$ sig.\ & med.\ $\hat\beta$ width \\
\midrule
Llama-3.1-70B & $0$ & $46$ & $73$ & $\mathbf{0.61}$ \\
Qwen3-32B     & $3$ & $73$ & $53$ & $1.06$ \\
gpt-oss-20b   & $7$ & $87$ & $50$ & $1.15$ \\
\bottomrule
\end{tabular}
\caption{Per-run bootstrap uncertainty (Experiment~E, $B{=}1000$; all values
  \%\ except the last). ``$\bar{\hat\beta}\!\le\!0$'' is runs with non-positive
  mean anchor gain. ``$\hat\beta$ CI $\ni 0$'' is agent-runs whose anchor CI
  contains zero. ``$\hat\alpha$ sig.'' is runs whose consensus-gain CI excludes
  zero. The last column is the median width of the per-agent $\hat\beta$ CI.}
\label{tab:expe}
\end{table}

\subsection{Per-Run Parameter Uncertainty}
\label{sec:uncertainty}
Block-bootstrapping each fit (Experiment~E $B{=}1000$ resamples of the
  transition pool) quantifies how well-determined the recovered parameters are
  within a single run (Table~\ref{tab:expe}). Two things follow. First, the
  model is not a perfect fit: the mean anchor gain is non-positive in $3\%$
  (Qwen) and $7\%$ (gpt-oss) of runs, and individual anchor confidence intervals (CIs) contain zero in
  $46\%$, $73\%$, and $87\%$ of agent-runs, the per-agent anchor is often
  not significant on its own. Second, and consistently with every other
  diagnostic, the  {uncertainty itself tracks the gradient}: the median
  anchor-CI width is $0.61$ for Llama against $1.06$ and $1.15$ for Qwen and
  gpt-oss, so Llama's anchors are roughly twice as well-determined. The
  consensus-gain CI width, by contrast, is essentially constant across families
  ($\approx0.32$). It is specifically the  {anchor} that is sharper where
  held-out validation certifies it. These per-run intervals are wide because a
  single run contributes only $K{=}5$ transitions, which is why we base the
  main claims on the population-level and held-out evidence above rather than on
  per-run significance. The bootstrap is reported here for the ordering it
  reveals, not as a per-agent test.

\paragraph{Anchor drift.}
We also tested whether the recovered anchor is stable  {within} a
  deliberation (Experiment~D): refitting on an early and a late window of rounds
  and comparing anchors. This diagnostic did not generalise, each
  window has too few transitions to recover a stable anchor, so the early/late
  comparison is dominated by fit noise rather than a consistent drift signal,
  and we therefore draw no conclusions from it.%
  % TODO(camera-ready): de-anonymise this URL (reveals author handle).
  % \footnote{Full negative
  % analysis: \url{https://github.com/apurbapokharel/BetterDeliberation/issues/4}.}

\paragraph{Anchor diagnostic accuracy.}
Anchor diagnostic accuracy is discussed in Appendix~\ref{app:accuracy}.

\section{Conclusion}
We modelled multi-agent LLM deliberation as a closed-loop dynamical system in
  which each agent carries a hidden per-agent anchor that pulls its opinion
  toward a latent prior independent of its neighbours. The model nests DeGroot
  ($\beta_i{=}0$) and Friedkin--Johnsen ($\mathbf{b}_i{=}\mathbf{x}_i(0)$) as
  special cases, and deliberation empirically settles within
  $\mathrm{conv}\{\mathbf{b}_i\}$ (\S\ref{sec:geometry}), reproducing the
  gold-class probability leaving the hull of initial opinions that linear
  consensus forbids (Property~\ref{prop:hull}, Eq.~\eqref{eq:escape}). Fitting the model
  by least squares and validating it on held-out seeds turns parameter recovery
  into an unsupervised model-selection test. On a symptom$\rightarrow$disease
  benchmark it certifies a transferable latent anchor for Llama-3.1-70B,
  reduces to the linear baselines for gpt-oss-20b, and places Qwen3-32B between
  them. Hidden-anchor behaviour is thus a spectrum across model families rather
  than a uniform property: the anchor gain is comparable across families,
  but only when the anchor sits far from the initial opinions does deliberation
  escape the hull and demand the full closed-loop model. Grounding the inferred
  anchor in model internals, and turning the open-loop contention schedule into
  a predictive controller of the dynamics, are the natural next steps.

\section*{Limitations}
We deliberately state the weaknesses of this work plainly.

\textbf{The positive result rests on one model.} Of the three families tested,
  held-out validation certifies a transferable latent anchor only for
  Llama-3.1-70B. For Qwen3-32B the margin over the linear baselines is within
  noise, and for gpt-oss-20b the baselines win outright. On this benchmark the
  ``hidden anchor'' is therefore close to a single-model phenomenon. Three
  open-weight models, on one English symptom$\rightarrow$disease task with
  $10$ cases, $n{=}3$ agents and $K{=}5$ rounds, might not license claims about
  model families in general.

\textbf{The anchor is weakly identified.} A single run identifies only the
  product $\boldsymbol{\gamma}_i=\beta_i\mathbf{b}_i$. The anchor
  $\mathbf{b}_i=\boldsymbol{\gamma}_i/\beta_i$ is recovered only by stacking
  multiple seeds, is ill-conditioned whenever $\hat\beta_i$ is small, and is
  further biased by the simplex projection. With only $K{=}5$ transitions per
  run, per-agent anchor confidence intervals contain zero in $46$--$87\%$ of
  agent-runs and the mean anchor gain is non-positive in up to $7\%$. The model
  is not significant at the per-run level. Every headline claim rests on
  population-level aggregation, and even there the held-out fit is modest: a
  mean $R^2(\Delta\mathbf{x})$ of $0.44$ for the best family (Llama) while the
  other two families sit at or below the linear baselines on held-out data.

\textbf{The anchor is inferred, not measured.} The anchor is a latent quantity
  fit to output-probability trajectories, not read from model internals. We do
  not show it corresponds to any actual internal representation, so reading
  $\mathbf{b}_i$ as the model's ``internal belief'' is an interpretation, not a
  verified claim. 
  % The linear discrete-time update is also a coarse
  % approximation of a strongly nonlinear generation process, and we do not
  % analyse the structure of the fit residuals.

\textbf{Escape is threshold-sensitive and orthogonal to accuracy.} Whether a
  trajectory escapes the initial hull depends on the margin required: at a
  negligible tolerance all three models escape in $\approx75\%$ of runs, and
  the family separation appears only at a hand-chosen $0.10$ margin. The
  dynamics we model are also orthogonal to correctness. The most dynamic family
  (Llama) is the least accurate ($43\%$), so the phenomenon we explain does not
  by itself improve deliberation. 
  % Finally, the contention-control application
  % is not part of this paper. Our attempt to verify it on this benchmark did not
  % produce a positive result, and we make no claim about controlling the
  % dynamics.

\section*{Writing Assistance}
We used Claude, an AI assistant, to help with writing in ways consistent with
  the ACL policy on AI assistance. Its use was limited to surface-level support:
  improving clarity, grammar, and phrasing of text we authored, and minor
  formatting and editing assistance. All research ideas, the model and its
  analysis, the experiments, and the claims are entirely our own, and we
  verified the correctness of all content. The assistant was not used to
  generate scientific content, results, or citations.

% \section*{Acknowledgments}
% % Unnumbered. Leave empty for anonymous review submission; fill for
% % camera-ready only.

\bibliography{custom}

\appendix

\section{Experimental Setup Details}
\label{app:setup}

\subsection{Models}
We use open-weight instruction-tuned LLMs only, for reproducibility.
The primary models are Llama-3.1-70B-Instruct~\cite{TODO-llama3} and
  Qwen3-32B~\cite{TODO-qwen3}, and we add one non-Llama,
  non-Qwen family (gpt-oss-20b~\cite{TODO-gptoss}) so that the
  generalisation claim is not specific to a single training lineage.
Llama and Qwen run at 4-bit quantisation via \texttt{bitsandbytes}
  (NF4 with double quantisation). gpt-oss-20b runs at its native MXFP4
  precision. Decoding uses temperature $0.7$, top-$p$ $0.9$, and a $512$-token
  cap per turn.

\subsection{Task and Dataset}
The task is symptom$\rightarrow$disease diagnosis: each agent is presented
  with a symptom set and asked for a ranked top-5 list of candidate diseases
  with self-reported probabilities, drawn from a $42$-class
  symptom--disease benchmark~\cite{TODO-symptom}. We use $10$ cases, each with
  a distinct gold disease, run with $3$ random seeds per case
  ($30$ trajectories per model, $90$ in total). A single deliberation domain
  is sufficient for the present claim, which concerns the per-family
   {dynamics} of deliberation rather than cross-domain generalisation.
  Extending the study to further domains (e.g.\ sentiment or legal-judgement
  classification) is left to future work.

\subsection{Prompt Templates}
\label{app:prompts}
All agents share the system message
\begin{quote}\small\ttfamily
Respond only in the requested format. Do not write your reasoning aloud.
\end{quote}
Round $0$ elicits each agent's initial opinion from the symptom set alone, with
  no neighbour context. Placeholders in braces are filled per case:
  \texttt{\{SYMPTOMS\}} is the case symptom string and
  \texttt{\{DISEASES\}} is the fixed list of the $42$ class labels.
\begin{quote}\small\ttfamily
You are given the following symptoms: \{SYMPTOMS\}.\\
You must choose ONLY from these diseases: \{DISEASES\}.\\[2pt]
Provide your top 5 disease predictions with confidence probabilities (must sum to 1.0).\\
Format each line exactly as: RANK. DISEASE\_NAME: PROBABILITY\\[2pt]
Example:\\
1. Diabetes: 0.40\\
2. Hypertension: 0.25\\
3. Hypothyroidism: 0.15\\
4. Heart attack: 0.12\\
5. Malaria: 0.08\\[2pt]
Your top 5 predictions:\\[2pt]
Then after the predictions explain in 2-3 sentences strictly why your top prediction is the most likely given the symptoms, and why the other candidates are less likely.\\[2pt]
Explanation:
\end{quote}
Each reflection round ($1\le k\le K$) gives the agent its own and its
  neighbour's previous response, in round-robin order.
  \texttt{\{PROBLEM\}} restates the diagnosis task,
  \texttt{\{OWN\_PREV\}} and \texttt{\{NEIGHBOUR\_PREV\}} are the two
  prior-round answers (predictions plus explanations).
\begin{quote}\small\ttfamily
Deliberation Topic: \{PROBLEM\}\\[2pt]
You previously predicted:\\
\{OWN\_PREV\}\\[2pt]
Another agent predicted:\\
\{NEIGHBOUR\_PREV\}\\[2pt]
You must choose ONLY from these diseases: \{DISEASES\}.\\[2pt]
Considering both predictions and their explanations, provide your updated top 5 disease predictions with confidence probabilities (must sum to 1.0).\\
Format each line exactly as: RANK. DISEASE\_NAME: PROBABILITY\\[2pt]
Your updated top 5 predictions:\\[2pt]
Then explain in 2-3 sentences why your top prediction is the most likely given the symptoms, and why the other candidates are less likely.\\[2pt]
Explanation:
\end{quote}
The fixed \texttt{RANK. DISEASE\_NAME: PROBABILITY} format is what the parser
  maps to the per-round opinion vector $\mathbf{x}_i(k)\in\Delta^{d-1}$ over the
  $d{=}42$ classes.

\subsection{Metrics}
We report four metrics so that the results section can stay terse.
\begin{itemize}
  \item $R^2(\Delta\mathbf{x})$~\eqref{eq:r2}: one-step prediction quality on
    stacked transitions.
  \item  {Held-out MSE}: mean-squared one-step prediction error under
    leave-one-run-out, capturing transfer of recovered parameters across
    runs.
  \item  {Bootstrap CIs}: $B=1000$ resamples of the transition pool
    yielding $95\%$ confidence intervals on $\hat\beta_i$ and $\hat{\mathbf{b}}_i$.
  \item  {Hull-containment rate}: fraction of runs whose final-round
    opinions all lie in $\mathrm{conv}\{\hat{\mathbf{b}}_j\}_j$ on every
    coordinate, measuring whether deliberation settles within the recovered
    anchor hull (\S\ref{sec:geometry}).
\end{itemize}

\subsection{Anchor Diagnostic Accuracy}
\label{app:accuracy}
For completeness we report whether the deliberated consensus lands on the gold
  diagnosis: the round-$K$ mean opinion has its argmax on the gold class in
  $43\%$ (Llama-3.1-70B), $57\%$ (Qwen3-32B), and $57\%$ (gpt-oss-20b) of runs.
  Notably the most dynamic family (Llama) is the  {least} accurate, so the
  rich, hull-escaping dynamics are not driven by movement toward the correct
  answer.
This is consistent with the scope of this work: we model and explain the
   {behaviour} of multi-agent deliberation, the latent-anchor dynamics
  and when they govern a model, not its task accuracy. Diagnostic
  correctness is orthogonal to the dynamical claim: a hidden anchor pulls an
  agent toward its own prior whether or not that prior is correct. Improving
  accuracy is a separate problem addressable by a richer deliberation
  framework (e.g.\ a judge agent or retrieval augmentation).

\section{Supplementary Tables}
\label{app:supp}
This appendix collects quantities that are cited in the body but do not warrant
  a table there. We add to it as further such numbers accumulate. Floats may be
  typeset away from this text. The tables in this section are, by clickable
  reference:
\begin{itemize}[leftmargin=1.4em,itemsep=1pt,topsep=2pt]
  \item Table~\ref{tab:app-range}: gold-coordinate per-run dispersion
    (\S\ref{sec:traj-obs}).
  \item Table~\ref{tab:app-selection}: per-disease held-out $R^2$ behind the
    model-selection counts (\S\ref{sec:heldout}).
\end{itemize}

\begin{table}[t]
\centering
\small
\begin{tabular}{lc}
\toprule
Model & Mean per-run range \\
\midrule
Llama-3.1-70B & $\mathbf{0.26}$ \\
Qwen3-32B     & $0.09$ \\
gpt-oss-20b   & $0.12$ \\
\bottomrule
\end{tabular}
\caption{Gold-coordinate dispersion within a deliberation
  (\S\ref{sec:traj-obs}). For each run the per-run range is the spread
  (maximum minus minimum) of the gold-class probability across rounds
  $0$--$K$. The table reports the mean over the $30$ runs per model. Llama's
  gold coordinate moves several times more than the other families, the
  quantitative form of the non-uniform trajectories in
  Figure~\ref{fig:dynamics_limits}.}
\label{tab:app-range}
\end{table}

\begin{table}[t]
\centering
\small
\begin{tabular}{lrrrc}
\toprule
Disease & DeGroot & FJ & Full & Best \\
\midrule
\multicolumn{5}{l}{\textit{Llama-3.1-70B}\quad(Full best in $8/10$)}\\
Fungal infection   & $0.02$ & $0.02$ & $\mathbf{0.73}$ & Full \\
Allergy            & $0.05$ & $0.02$ & $\mathbf{0.66}$ & Full \\
GERD               & $0.08$ & $0.07$ & $\mathbf{0.24}$ & Full \\
Chronic chol.      & $\mathbf{0.07}$ & $0.06$ & $-0.03$ & DeGroot \\
Drug reaction      & $0.13$ & $0.14$ & $\mathbf{0.55}$ & Full \\
COVID              & $0.00$ & $0.00$ & $\mathbf{0.81}$ & Full \\
Peptic ulcer       & $0.05$ & $0.05$ & $\mathbf{0.74}$ & Full \\
AIDS               & $0.12$ & $0.08$ & $\mathbf{0.49}$ & Full \\
Diabetes           & $0.00$ & $0.00$ & $\mathbf{0.62}$ & Full \\
Gastroenteritis    & $\mathbf{-0.02}$ & $-0.05$ & $-0.40$ & DeGroot \\
\midrule
\multicolumn{5}{l}{\textit{Qwen3-32B}\quad(Full best in $7/10$)}\\
Fungal infection   & $0.26$ & $0.26$ & $\mathbf{0.34}$ & Full \\
Allergy            & $0.24$ & $\mathbf{0.24}$ & $-0.06$ & FJ \\
GERD               & $0.03$ & $0.04$ & $\mathbf{0.08}$ & Full \\
Chronic chol.      & $0.13$ & $0.13$ & $\mathbf{0.24}$ & Full \\
Drug reaction      & $0.07$ & $0.06$ & $\mathbf{0.25}$ & Full \\
COVID              & $\mathbf{0.00}$ & $-0.01$ & $-0.19$ & DeGroot \\
Peptic ulcer       & $0.04$ & $0.04$ & $\mathbf{0.10}$ & Full \\
AIDS               & $\mathbf{0.08}$ & $0.07$ & $-0.25$ & DeGroot \\
Diabetes           & $0.09$ & $0.09$ & $\mathbf{0.13}$ & Full \\
Gastroenteritis    & $0.10$ & $0.08$ & $\mathbf{0.17}$ & Full \\
\midrule
\multicolumn{5}{l}{\textit{gpt-oss-20b}\quad(Full best in $2/10$)}\\
Fungal infection   & $\mathbf{0.18}$ & $0.17$ & $-0.12$ & DeGroot \\
Allergy            & $0.16$ & $\mathbf{0.16}$ & $0.08$ & FJ \\
GERD               & $0.16$ & $0.16$ & $\mathbf{0.22}$ & Full \\
Chronic chol.      & $\mathbf{0.13}$ & $0.13$ & $0.08$ & DeGroot \\
Drug reaction      & $\mathbf{0.10}$ & $0.09$ & $0.06$ & DeGroot \\
COVID              & $\mathbf{0.02}$ & $-0.01$ & $-9.43$ & DeGroot \\
Peptic ulcer       & $0.22$ & $0.18$ & $\mathbf{0.35}$ & Full \\
AIDS               & $\mathbf{0.06}$ & $0.05$ & $-0.79$ & DeGroot \\
Diabetes           & $\mathbf{0.21}$ & $0.20$ & $0.13$ & DeGroot \\
Gastroenteritis    & $\mathbf{0.04}$ & $0.02$ & $0.01$ & DeGroot \\
\bottomrule
\end{tabular}
\caption{Per-disease held-out $R^2(\Delta\mathbf{x})$ behind the
  model-selection counts of \S\ref{sec:heldout} and the ``Full sel.\ \%''
  column of Table~\ref{tab:nested}. For each of the $10$ disease groups we fit
  the three nested models (DeGroot, Friedkin--Johnsen, Full) on two seeds and
  score one-step prediction on the held-out seed, averaged over the three
  leave-one-seed-out folds. The best model per row is in bold, and the
  ``Best'' column names it. Counting the Full-best rows gives the selection
  counts ($8/10$, $7/10$, $2/10$). The single extreme gpt-oss COVID entry
  ($-9.43$) is the catastrophic overfit that drags the gpt-oss mean held-out
  $R^2$ to $-0.94$ in Table~\ref{tab:nested}. Values are read from
  \texttt{heldout\_nested.csv} produced by \texttt{heldout\_nested.py}.}
\label{tab:app-selection}
\end{table}

\section{Reproducibility}
\label{app:repro}
% TODO(camera-ready): replace with the public repository link.
All code, the stored deliberation trajectories, and the analysis scripts that
  regenerate every table and figure are included in the supplementary material.

\paragraph{Environment.}
Experiments run under Python~3 with NumPy and SciPy for the analysis and
  PyTorch with Hugging Face Transformers for deliberation. The three models,
  their quantisation settings, and the decoding hyperparameters are listed in
  Appendix~\ref{app:setup}. A pinned dependency list is included in the
  repository.

\paragraph{Deliberation.}
Each $(\text{model},\text{case})$ cell is run with the protocol of
  \S\ref{sec:protocol}: $n{=}3$ agents on a directed ring, $K{=}5$ reflection
  rounds, and the $3$ fixed random seeds. The deliberation step writes one JSON
  trajectory per run, recording every agent's per-round probability vector over
  the $42$ classes. These stored trajectories are the sole input to the
  analysis. No model has to be re-run to reproduce any number in the paper.

\paragraph{Analysis.}
The analysis is deterministic: it reads the stored trajectories and refits the
  nested models by ordinary least squares (\S\ref{sec:sysid}). The only
  stochastic step, the Experiment~E block bootstrap, is seeded ($B{=}1000$,
  fixed seed), so its confidence intervals reproduce exactly. Each table and
  figure maps to a single script:
\begin{itemize}[leftmargin=1.4em,itemsep=1pt,topsep=2pt]
  \item Table~\ref{tab:nested}: in-sample fits and the nested-model comparison,
    with the leave-one-seed-out held-out block and the per-disease selection
    counts (the latter detailed in Table~\ref{tab:app-selection}).
  \item Table~\ref{tab:geometry}: recovered-anchor strength and overshoot past
    the initial band, with the steady-state containment rate.
  \item Table~\ref{tab:escape}: the gold-coordinate escape margin beyond
    $\mathrm{conv}\{\mathbf{x}_j(0)\}$.
  \item Table~\ref{tab:expe}: the per-run bootstrap confidence intervals of
    Experiment~E.
  \item Table~\ref{tab:app-range}: the per-run gold-coordinate dispersion.
  \item Table~\ref{tab:app-heldout} and the figures of
    Appendix~\ref{app:gallery}: the per-run held-out validation and the native
    per-run plots for the showcased Llama COVID case.
\end{itemize}
Every reported value was regenerated from the released trajectories with these
  scripts, and the script names and exact commands are given in the repository
  README.

\section{Per-Run Figure Gallery}
\label{app:gallery}
\begin{figure}[t]
\centering
\includegraphics[width=\linewidth]{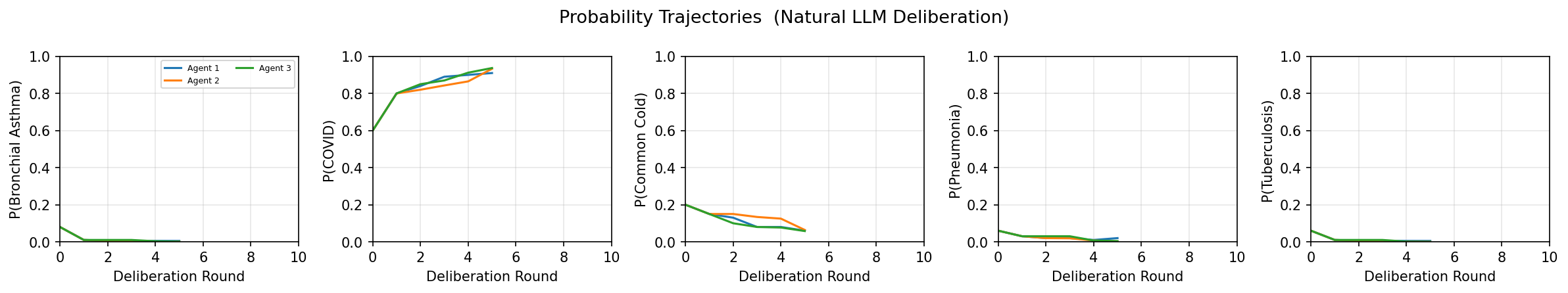}
\caption{Natural LLM deliberation (Experiment~B), per-class probability over
  rounds for all three agents. The gold class (COVID) climbs from $0.60$ to
  $0.93$, above every agent's round-$0$ value, so the trajectory leaves
  $\mathrm{conv}\{\mathbf{x}_j(0)\}$ on that coordinate. This is the escape
  that the open-loop rules forbid (Property~\ref{prop:hull}).}
\label{fig:app-traj}
\end{figure}

\begin{figure}[t]
\centering
\includegraphics[width=\linewidth]{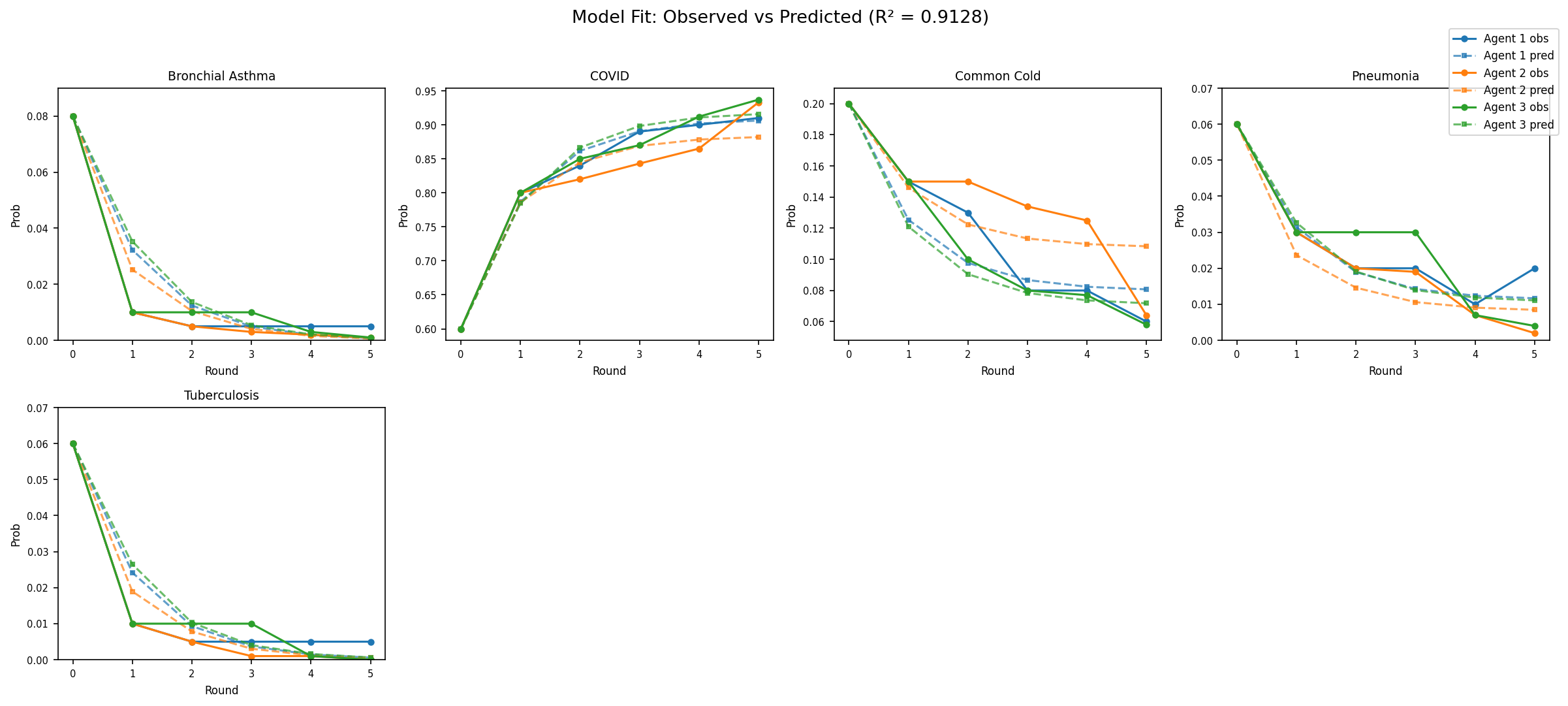}
\caption{Hidden-anchor system identification (Experiment~C,
  \S\ref{sec:sysid}) for the same run. Observed (solid) against the one-step
  model prediction (dashed) per class, with in-sample
  $R^2(\Delta\mathbf{x})=0.91$.}
\label{fig:app-fit}
\end{figure}

\begin{figure}[t]
\centering
\includegraphics[width=\linewidth]{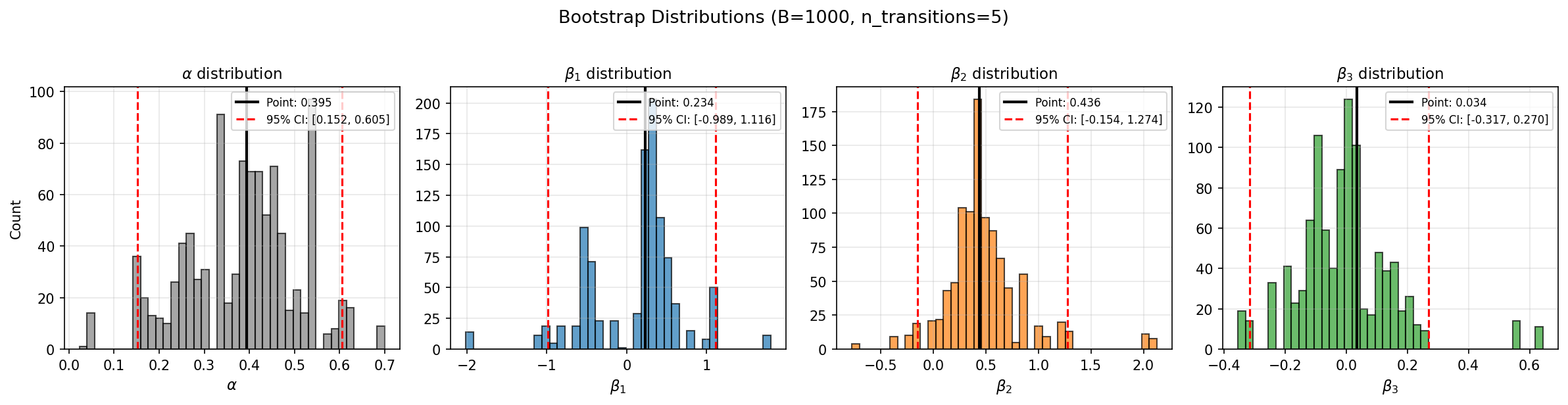}
\caption{Per-run bootstrap distributions (Experiment~E,
  \S\ref{sec:uncertainty}; $B{=}1000$, five transitions) of the recovered
  parameters for the same run. Companion to Figure~\ref{fig:app-ci-forest}.}
\label{fig:app-ci-dist}
\end{figure}

\begin{figure}[t]
\centering
\includegraphics[width=\linewidth]{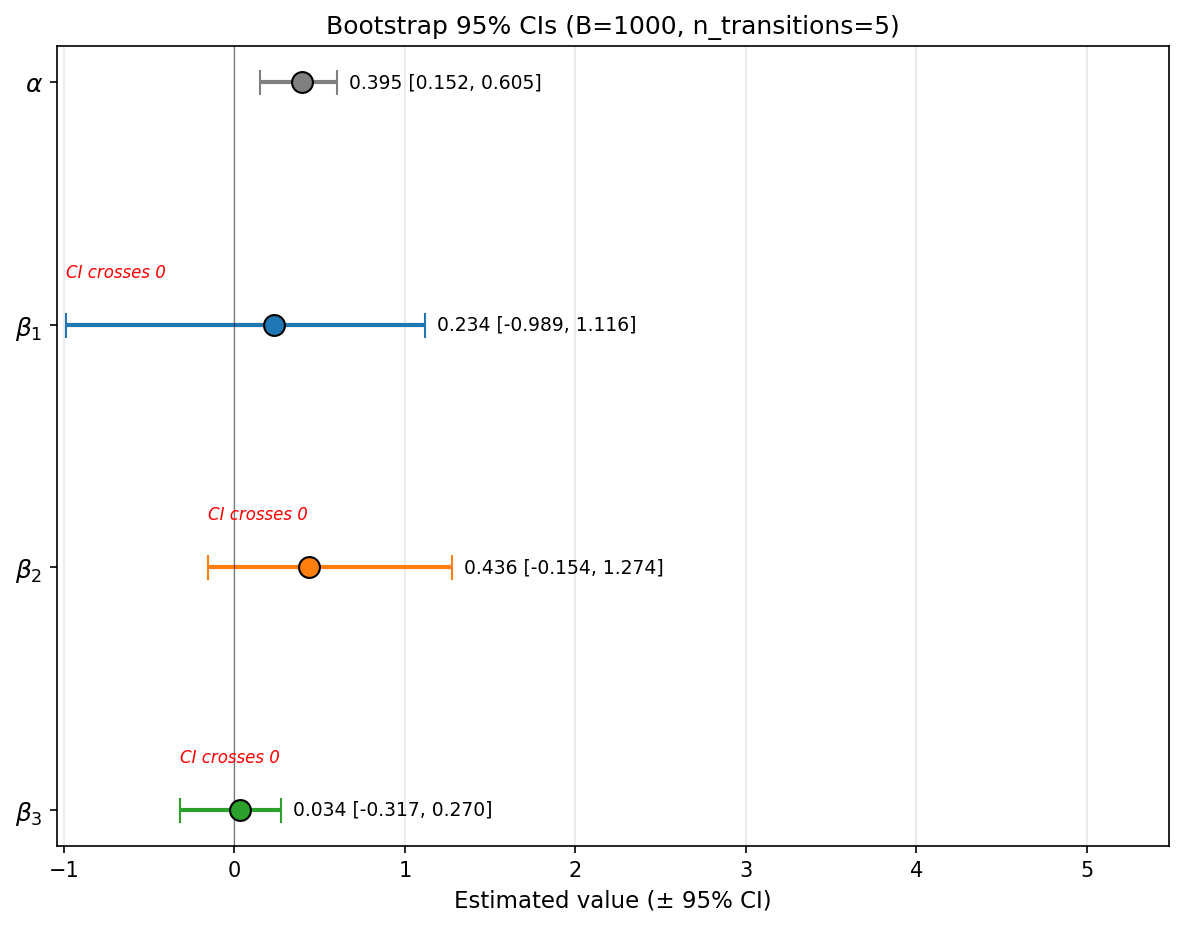}
\caption{Per-run $95\%$ confidence intervals (Experiment~E,
  \S\ref{sec:uncertainty}) from the bootstrap of
  Figure~\ref{fig:app-ci-dist}. The consensus gain $\hat\alpha$ excludes zero,
  while every $\hat\beta_i$ interval crosses zero: with only five transitions
  per run the individual anchor strengths are not separately identifiable,
  which is why the anchor signal is read at the population level rather than
  per run.}
\label{fig:app-ci-forest}
\end{figure}

For concreteness we walk one representative run through the analysis battery
  of \S\ref{sec:procedures}. We use Llama-3.1-70B on the COVID case (seed~2),
  the family with the strongest hidden-anchor signal. These are the native
  per-run outputs of Experiments~B, C, E and F. We omit Experiment~A here,
  since this run starts from near-identical agent beliefs, so the open-loop
  baseline is static and uninformative as a picture. The qualitative behaviour
  shown is representative of the Llama runs that the aggregate tables
  summarise. This section is placed last so that its figures do not interrupt
  the surrounding appendix text. Floats may be typeset away from this text. The
  figures and table in this section are, by clickable reference:
\begin{itemize}[leftmargin=1.4em,itemsep=1pt,topsep=2pt]
  \item Figure~\ref{fig:app-traj}: natural deliberation trajectory
    (Experiment~B).
  \item Figure~\ref{fig:app-fit}: hidden-anchor system-identification fit
    (Experiment~C).
  \item Figure~\ref{fig:app-ci-dist}: per-run bootstrap distributions
    (Experiment~E).
  \item Figure~\ref{fig:app-ci-forest}: per-run parameter confidence intervals
    (Experiment~E).
  \item Table~\ref{tab:app-heldout}: held-out leave-one-seed-out validation
    (Experiment~F).
\end{itemize}

\begin{table}[t]
\centering
\small
\begin{tabular}{lccc}
\toprule
Held-out seed & $R^2(\Delta\mathbf{x})$ & $R^2(\text{state})$ & MSE \\
\midrule
1 (vs.\ 2,\,3) & $0.22$  & $0.99$ & $0.0011$ \\
2 (vs.\ 1,\,3) & $0.18$  & $0.98$ & $0.0019$ \\
3 (vs.\ 1,\,2) & $-0.76$ & $0.99$ & $0.0013$ \\
\midrule
Mean           & $-0.12$ & $0.99$ & $0.0014$ \\
\bottomrule
\end{tabular}
\caption{Held-out leave-one-seed-out validation (Experiment~F,
  \S\ref{sec:heldout}) for the Llama COVID case. The three deliberation seeds
  are split into two training seeds and one held-out seed. We fit the
  hidden-anchor model (\S\ref{sec:sysid}) on the pooled one-step transitions
  of the two training seeds (the ensemble fit), forward-simulate from the
  held-out seed's round-$0$ opinions, and score the prediction against that
  seed's observed trajectory. $R^2(\Delta\mathbf{x})$ is the one-step
  prediction quality on stacked transitions. $R^2(\text{state})$ is the
  trajectory-level fit of the simulated states. MSE is the mean squared
  per-coordinate error. The trajectory-level fit is high and stable
  ($R^2(\text{state})\ge0.98$) while the one-step $R^2(\Delta\mathbf{x})$ is
  noisy and turns negative on seed~3, an artefact of having only five
  transitions per seed. For reference the in-sample $R^2(\Delta\mathbf{x})$
  averages $0.82$. Numbers are read from \texttt{held\_out\_results.json}
  produced by \texttt{run\_experiment\_f}.}
\label{tab:app-heldout}
\end{table}

\end{document}